\documentclass[letterpaper]{article}
\usepackage[preprint]{aaai2027}
\usepackage[hyphens]{url}
\usepackage{natbib}
\usepackage{amsmath,amssymb,amsthm}
\usepackage{mathtools}
\usepackage{booktabs}
\usepackage{multirow}
\usepackage{colortbl}
\usepackage[table]{xcolor}
\usepackage{graphicx}
\usepackage{float}
\usepackage{caption}
\usepackage{algorithm}
\usepackage{algorithmic}
\newtheorem{proposition}{Proposition}

\newcommand{\stitle}[1]{\vspace*{0.4em}\noindent{\bf #1.\/}}
\begin{document}

\title{OPDSearch+: On-Policy Distillation with RL Refinement \\for Search-Augmented Reasoning}
\author{
    Qinglin Ye\textsuperscript{\rm 1,2,*},
    Zhiyuan Gu\textsuperscript{\rm 1,3,*},
    Jingjie Xia\textsuperscript{\rm 1,2,*},
    Yiheng Zhang\textsuperscript{\rm 4},
    Kaiyan Zhao\textsuperscript{\rm 5},
    Shunchao Zheng\textsuperscript{\rm 6},\\
    Yuhang Mu\textsuperscript{\rm 7},
    Wenchao Du\textsuperscript{\rm 1},
    Yiming Wang\textsuperscript{\rm 8,\dag}
}
\affiliations{
    \textsuperscript{\rm 1}University of Chinese Academy of Sciences\quad
    \textsuperscript{\rm 2}Institute of Computing Technology, Chinese Academy of Sciences\quad
    \textsuperscript{\rm 3}Institute of Automation, Chinese Academy of Sciences\\
    \textsuperscript{\rm 4}University of Macau\quad
    \textsuperscript{\rm 5}Wuhan University\quad
    \textsuperscript{\rm 6}Georgia Institute of Technology\quad
    \textsuperscript{\rm 7}Northwestern Polytechnical University\quad
    \textsuperscript{\rm 8}University of Hong Kong\\
    \textsuperscript{\rm *}Equal contribution.\quad
    \textsuperscript{\rm \dag}Corresponding author.
}
\maketitle

\begin{abstract}
Search-augmented reasoning remains difficult for small language models. On-policy distillation (OPD) from trained teachers offers a promising direction, but suffers from two issues: (1) high-quality multi-turn search trajectories depend on dynamic retriever responses, making SFT data prohibitively expensive to collect at scale; (2) task-specifically trained teachers incur substantial training cost, while directly applying OPD with an off-the-shelf teacher without task-specific fine-tuning constrains the student to the teacher's performance ceiling and suffers from severe training instability.
We propose \textbf{OPDSearch+}, the first distillation paradigm that requires no teacher fine-tuning for search-augmented reasoning. We investigate the role of a frozen off-the-shelf instruct model as the teacher in on-policy distillation, and reveal a key insight: \emph{the teacher reshapes the student's policy distribution so that subsequent RL converges to a superior solution that RL alone cannot reach}. In stage one, the student interacts with a live search engine and is distilled via a per-position forward KL objective, transferring reasoning decomposition and evidence integration skills without any task-specific teacher training. In stage two, RL refines the distilled student from a richer behavioral foundation, achieving performance that RL alone cannot reach from scratch. Across seven QA benchmarks, OPDSearch+ with a 3B model consistently outperforms all prior 3B RL baselines, achieving gains of $13.1\%$ on HotpotQA and $8.5\%$ on 2WikiMultihopQA.
\end{abstract}

% ====================================================================
\section{Introduction}
% ====================================================================

\begin{figure}[t]
\centering
\includegraphics[width=\columnwidth]{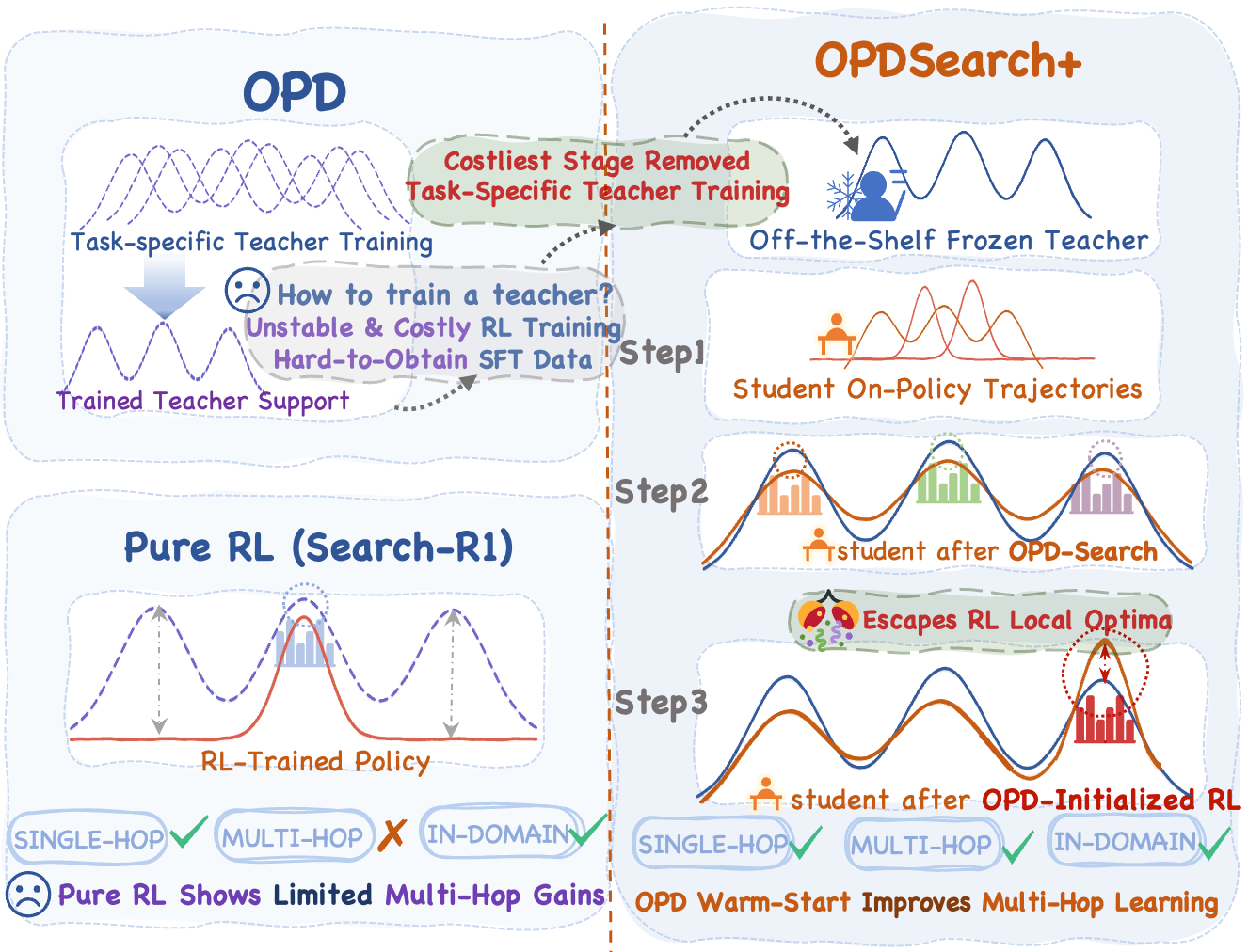}
\caption{Left: Prior OPD requires costly task-specific teacher training; pure RL shows limited multi-hop gains. Right: OPDSearch+ eliminates teacher training by using a frozen off-the-shelf teacher. The teacher reshapes the student's distribution via on-policy distillation, enabling subsequent RL to escape local optima and achieve strong single-hop, multi-hop, and in-domain performance.}
\label{fig:intro}
\end{figure}

\begin{figure*}[t]
\centering
\includegraphics[width=\textwidth]{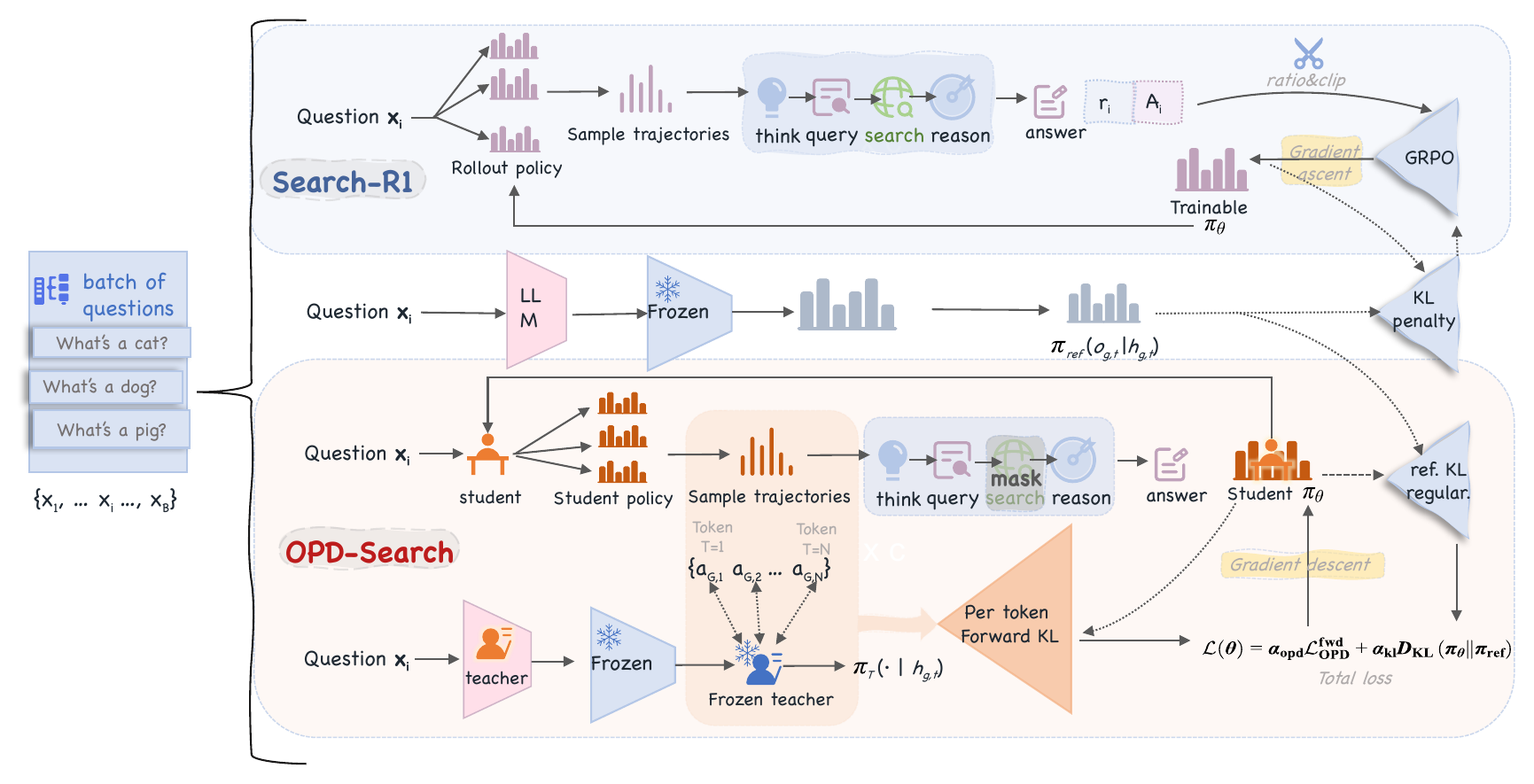}
\caption{Comparison between RL-based search agent training (Search-R1) and our On-Policy Distillation (OPDSearch+) framework. (a) Search-R1 uses outcome reward (EM/F1) to train the model via GRPO, which conflates retrieval quality with answer correctness and suffers from reward hacking. (b) OPDSearch+ generates on-policy trajectories from the student interacting with a live search engine, then distills the teacher's token-level distribution onto these trajectories. The teacher's search behavior serves as implicit supervision for both reasoning and retrieval, providing a strong initialization for subsequent RL refinement.}
\label{fig:overview}
\end{figure*}

% ---------- Para 1: Background & Challenge ----------
Training LLMs to interact with search engines for knowledge-intensive QA has become a central research direction~\cite{schick2023toolformer,yao2023react,nakano2021webgpt}. On-policy distillation (OPD) from trained teachers offers a promising paradigm for transferring search-and-reason capabilities from larger models to smaller ones. However, existing OPD methods face two fundamental obstacles. First, high-quality multi-turn search trajectories depend on dynamic retriever responses, making SFT data prohibitively expensive to collect at scale, and unlike static text generation, these trajectories cannot be reused across different retrieval systems or corpus versions. Second, task-specifically trained teachers incur substantial training cost and instability, while directly applying OPD with an off-the-shelf teacher without task-specific fine-tuning constrains the student to the teacher's performance ceiling and suffers from severe training instability due to distributional misalignment between the teacher and the student's on-policy trajectories.

% ---------- Para 2: Our approach ----------
We propose \textbf{OPDSearch+}, a distillation paradigm that requires \textit{no teacher fine-tuning} and resolves both obstacles. Our key insight is that \emph{the teacher's role is not to provide a performance ceiling for imitation, but to reshape the student's policy distribution so that subsequent RL converges to a superior solution that RL alone cannot reach}. Concretely, a frozen off-the-shelf instruct model serves as the teacher: it provides token-level supervision via a per-position forward KL objective on student-generated trajectories, transferring reasoning decomposition and evidence integration capabilities without any task-specific teacher training. The student interacts with a live search engine, and the teacher evaluates these on-policy trajectories, avoiding both the data construction challenge and the teacher training cost. Once distillation has expanded the student's expressive capacity, RL refines the distilled student from this richer behavioral foundation, achieving performance that RL alone cannot reach from scratch.

% ---------- Para 3: Results ----------
OPDSearch+ achieves mean EM $0.4402$ across seven QA benchmarks, exceeding all 3B RL baselines including the previous best GiGPO-Instruct at $0.421$. Gains are particularly pronounced on multi-hop tasks: $+13.1\%$ on HotpotQA and $+8.5\%$ on 2Wiki over the best 3B baseline AutoRefine-Base.

% ---------- Para 4: Contributions ----------
Our contributions are listed as follows:
\begin{itemize}
    \item We propose OPDSearch+, the first on-policy distillation framework for interactive retrieval that uses a frozen off-the-shelf teacher, providing token-level supervision on live search trajectories without costly data construction or task-specific teacher training.

    \item We show that OPD reshapes the student policy into a stronger initialization for RL, enabling it to outperform both the teacher and RL from the same base model. Theoretically, we establish that clipped forward KL controls gradient variance under teacher–student mismatch; empirically, it remains more stable than reverse-KL variants while increasing policy entropy, broadening behavioral diversity, and reducing search turns.

    \item The resulting two-stage pipeline achieves mean EM $0.4402$ across seven QA benchmarks with a 3B student, outperforming all 3B baselines (the best prior result is $0.421$). Multi-hop gains are particularly pronounced: $+13.1\%$ on HotpotQA and $+8.5\%$ on 2WikiMultihopQA, demonstrating that distillation transfers compositional reasoning more efficiently than RL discovers it from scratch.
\end{itemize}

% ====================================================================
\section{Related Work}
% ====================================================================

\stitle{RL for Search-Augmented Reasoning}
Search-R1~\cite{jin2025searchr1} demonstrates that LLMs can learn to interleave reasoning with retrieval via RL with outcome reward. Subsequent work including R1-Searcher~\cite{r1searcher}, ReSearch~\cite{research}, and ASearcher~\cite{asearcher} extends this to diverse settings but retains the outcome-only reward paradigm. Process-reward approaches such as StepSearch~\cite{wang2025stepsearch} and GiGPO~\cite{feng2025gigpo} add step-level supervision but require additional engineering (GPT-generated sub-questions, contrastive group construction). All RL-based methods share fundamental challenges with sample efficiency and training stability, particularly for smaller models.

\stitle{Knowledge Distillation for Reasoning LLMs}
Knowledgedistillation~\cite{hinton2015distilling} has been widely applied to compress LLMs while preserving reasoning capabilities. Offline approaches~\cite{kim2016sequence} train on teacher-generated data but suffer from train-test distribution mismatch: the student sees teacher trajectories during training but must follow its own distribution at inference, leading to error accumulation. On-policy distillation (OPD)~\cite{agarwal2024onpolicy} addresses this by scoring student-generated samples with the teacher, typically using reverse KL minimization. Recent work explores the interplay between KL direction and training stability: EOPD~\cite{jin2026eopd} uses entropy-adaptive KL mixing, Decoupled-KL~\cite{zhao2026decoupled} analyzes the design space of prefix source and KL direction and finds that forward KL is essential for preventing entropy collapse in long-sequence distillation, and KDRL~\cite{xu2025kdrl} jointly optimizes KL distillation with RL objectives. Recently, several works have begun extending distillation to interactive environments: SD-Search~\cite{sdsearch} performs on-policy self-distillation within a search-augmented reasoning loop via JSD minimization at search-query positions (using the same model as both teacher and student); Agent Distillation~\cite{agentdistill} distills full agent trajectories including retrieval and code tool calls; TT-OPD~\cite{ttopd} applies turn-level truncated OPD in multi-tool interactive settings. However, these methods either rely on self-distillation (unable to leverage stronger models' capabilities), require task-specifically trained teachers, or use mode-seeking objectives (JSD/reverse KL). In contrast, OPDSearch+ uses a \textbf{frozen off-the-shelf instruct model} (requiring no task-specific fine-tuning) as a cross-model teacher and employs a \textbf{per-position forward KL} objective (a clipped importance-weighted estimator on student prefixes) for distillation in an interactive retrieval environment---this combination encourages the student to both leverage the large model's generalization capabilities and preserve diversity of search strategies, without any expensive teacher training pipeline.

\stitle{KD-RL Hybrid Methods}
Recent methods have explored combining KD with RL for reasoning tasks: RLAD~\cite{zhang2026rlad} uses advantage-weighted trust-region distillation, SPOT~\cite{lin2026spot} uses proximal on-policy distillation as RL initialization, TGPO~\cite{liu2026tgpo} has the teacher generate in the student's context, and SC-GRPO~\cite{shan2026scgrpo} uses self-conditioned KL as credit-assignment weights. These methods all operate in static text-generation settings (e.g., mathematical reasoning) where the model does not interact with external systems. In contrast, OPDSearch+ addresses the distinct challenge of \emph{interactive retrieval environments}, where trajectories depend on dynamic search engine responses and the teacher must provide guidance conditioned on live environment feedback.

% ====================================================================
\section{Method}
% ====================================================================

\subsection{Why On-Policy Distillation Suits Search Agents}
\label{sec:why_opd}

We establish three formal properties that justify why forward-KL on-policy distillation is particularly well-suited to search-augmented reasoning agents.

\begin{proposition}[Implicit multi-level supervision]
\label{prop:decomposition}
Let a trajectory $o=(o_1,\ldots,o_T)$ be partitioned into reasoning tokens $\mathcal{T}_R$, query tokens $\mathcal{T}_Q$, and answer tokens $\mathcal{T}_A$ with $\mathcal{T}_R \cup \mathcal{T}_Q \cup \mathcal{T}_A = \mathcal{M}$. Define the teacher-to-student importance ratio $r_t = \pi_{\text{tea}}(o_t\mid o_{<t}) / \pi_\theta(o_t\mid o_{<t})$ and its clipped version $\bar r_t=\operatorname{clip}(r_t,\epsilon,R_{\text{max}})$. Then the implemented forward-KL gradient decomposes as:
\begin{equation}
\nabla_\theta \mathcal{L}_{\text{OPD}}^{\text{fwd}} = -\frac{1}{|\mathcal{M}|}\left(\sum_{t\in\mathcal{T}_R} \bar r_t\, g_t + \sum_{t\in\mathcal{T}_Q} \bar r_t\, g_t + \sum_{t\in\mathcal{T}_A} \bar r_t\, g_t\right),
\end{equation}
where $g_t = \nabla_\theta \log\pi_\theta(o_t\mid o_{<t})$. Consequently, for query tokens where the teacher assigns high probability but the student does not, a raw ratio $r_t>1$ amplifies the gradient through $\bar r_t$, up to the cap $R_{\text{max}}$---providing implicit query-quality supervision without a dedicated retrieval reward.
\end{proposition}

The decomposition follows from linearity of summation over token positions. The key consequence is that the clipped ratio $\bar r_t$ acts as a \emph{per-token adaptive reward}: tokens the teacher strongly endorses receive amplified gradients, up to $R_{\text{max}}$, regardless of their functional role in the trajectory. A well-formed entity-specific query can receive a large raw ratio $r_t$ when the teacher favors it but the student under-assigns probability, while a verbatim question-copy receives $r_t \approx 1$. This provides implicit reward shaping that outcome-based RL cannot achieve---outcome rewards assign identical credit to all tokens in the trajectory.

\begin{proposition}[Second-moment control under distributional mismatch]
\label{prop:variance}
Let $g_t = \nabla_\theta \log\pi_\theta(o_t\mid o_{<t})$ and let $\bar r_t=\operatorname{clip}(r_t,\epsilon,R_{\text{max}})$ be the detached clipped importance ratio. Then the per-token gradient second moment under the implemented forward-KL objective satisfies:
\begin{equation}
\mathbb{E}_{o_t \sim \pi_\theta}\!\left[\|\bar r_t\, g_t\|^2\right] \;\leq\; R_{\text{max}}^2 \cdot \mathbb{E}_{o_t \sim \pi_\theta}\!\left[\|g_t\|^2\right].
\end{equation}
In contrast, for the reverse-KL surrogate, the effective weight $c_t = \pi_\theta(o_t)/\pi_{\text{tea}}(o_t) - 1$ is unbounded and correlates positively with the sampling probability $\pi_\theta(o_t)$, so high-variance gradient terms are encountered frequently rather than rarely.
\end{proposition}

Since $\bar r_t \in [\epsilon, R_\text{max}]$ and is treated as stop-gradient, the bound follows from $\|\bar r_t\, g_t\|^2 \leq R_\text{max}^2\|g_t\|^2$ pointwise. The critical asymmetry is: under forward-KL, tokens with a large raw ratio $r_t$ receive a capped coefficient $\bar r_t\leq R_\text{max}$ and are those where $\pi_\theta$ is small---precisely the tokens \emph{rarely sampled} on-policy, so their high-magnitude contributions appear infrequently in minibatches. Under reverse-KL, the coefficient $c_t = \pi_\theta(o_t)/\pi_\text{tea}(o_t) - 1$ grows large when the student over-assigns probability relative to the teacher, and these tokens are sampled \emph{frequently} (proportional to $\pi_\theta$), creating a positive feedback loop that drives entropy collapse. This explains the empirical instability of reverse-KL variants observed in Section~\ref{sec:kl_comparison}.

\begin{proposition}[Distribution-shift bound for on-policy evaluation]
\label{prop:onpolicy}
Let $P_\theta^{\mathcal{R}}$ denote the prefix distribution induced by the student interacting with retriever $\mathcal{R}$, and $P_{\text{offline}}$ the distribution of pre-collected offline trajectories. Define the per-prefix forward KL $f(o_{<t}) = D_{\text{KL}}\bigl(\pi_{\text{tea}}(\cdot\mid o_{<t})\,\|\,\pi_\theta(\cdot\mid o_{<t})\bigr)$ bounded by $C$. Then:
\begin{equation}
\left|\mathbb{E}_{P_\theta^{\mathcal{R}}}[f] - \mathbb{E}_{P_{\text{offline}}}[f]\right| \;\leq\; C \cdot D_{\text{TV}}\!\left(P_\theta^{\mathcal{R}},\, P_{\text{offline}}\right).
\end{equation}
On-policy distillation sets $P_{\text{offline}} = P_\theta^{\mathcal{R}}$, eliminating the distribution-shift gap entirely.
\end{proposition}

The bound follows from treating the per-prefix KL as a bounded test function and applying the variational characterization of total variation distance. In search-augmented settings, the prefix includes retrieval results that depend on prior queries issued by $\pi_\theta$: even small policy changes alter which passages are retrieved, cascading through subsequent reasoning steps. Offline methods, which train on teacher-generated trajectories, suffer from this compounding distribution shift---the student sees teacher retrieval contexts during training but must follow its own at inference. On-policy distillation avoids this entirely by evaluating the teacher on the student's \emph{actual} retrieval contexts, ensuring the gradient always reflects the student's real operating conditions.

\subsection{Problem Setup}

We consider the search-augmented QA task following the Search-R1 framework~\cite{jin2025searchr1}. Given a question $x$, the model generates a multi-turn trajectory consisting of interleaved reasoning (\texttt{<think>}), search queries (\texttt{<search>}), retrieved passages (\texttt{<information>}), and a final answer (\texttt{<answer>}). The search engine $\mathcal{R}$ returns relevant passages for each query. A trajectory $o = (o_1, o_2, \ldots, o_T)$ is the full sequence of tokens generated by the model (reasoning, queries, answers), where retrieved passages are provided by the environment and not generated by the model.

\subsection{On-Policy Distillation for Search (OPDSearch+)}

\paragraph{Overview.}
OPDSearch+ trains a student policy $\pi_\theta$ (Qwen2.5-3B) to imitate a frozen teacher $\pi_\text{tea}$ (Qwen2.5-14B-Instruct) \emph{on the student's own search trajectories}. The key distinction from standard text-only OPD is that trajectories are generated through interaction with a live retrieval environment: the student issues search queries, receives real passages from the Wikipedia corpus, and must integrate this evidence. The teacher evaluates the student's complete trajectory (including the environmental feedback) and provides token-level supervision.

\paragraph{Training procedure.}
For each training batch of questions $\{x_1, \ldots, x_B\}$:
\begin{enumerate}
    \item \textbf{On-policy rollout}: The student $\pi_\theta$ generates $G$ trajectories per question by interacting with the search engine $\mathcal{R}$. Each trajectory $o_i^{(g)}$ includes the student's reasoning, queries, and the search engine's responses.
    \item \textbf{Teacher scoring}: The frozen teacher $\pi_\text{tea}$ computes token-level log-probabilities on the student-generated trajectories (excluding retrieved-passage tokens, which are environment-provided).
    \item \textbf{Student update}: The student parameters are updated to minimize the KL divergence between the student and teacher distributions on these trajectories.
\end{enumerate}

\paragraph{Loss function.}
The training objective combines two terms:
\begin{equation}
\mathcal{L}(\theta) = \alpha_\text{opd} \cdot \mathcal{L}_\text{OPD}(\theta) + \alpha_\text{kl} \cdot \mathcal{L}_\text{KL}(\theta),
\label{eq:total_loss}
\end{equation}
where $\mathcal{L}_\text{OPD}$ is the forward-KL distillation loss (a clipped importance-weighted estimator; see below) and $\mathcal{L}_\text{KL}$ is a KL regularization term against a frozen reference policy (the initial student checkpoint) to prevent catastrophic deviation.

\paragraph{Teacher-weighted cross-entropy (forward KL motivation).}
Our distillation objective minimizes the \emph{per-position forward KL} $D_\text{KL}(\pi_\text{tea}\|\pi_\theta)$ at each token position, estimated via importance weighting on student-generated prefixes. Let $\ell_t = \log\pi_\theta(o_t | o_{<t}, x; \mathcal{R})$ and $\ell_t^\text{tea} = \log\pi_\text{tea}(o_t | o_{<t}, x; \mathcal{R})$, and define the teacher-to-student ratio and its clipped version as $r_t = \exp(\ell_t^\text{tea} - \ell_t)$ and $\bar r_t=\operatorname{clip}(r_t,\epsilon,R_\text{max})$, respectively. The implemented loss is:
\begin{equation}
\mathcal{L}_\text{OPD}^{\text{fwd}}(\theta) = -\frac{1}{|\mathcal{M}|}\sum_{t \in \mathcal{M}} \text{sg}(\bar r_t) \cdot \log\pi_\theta(o_t | o_{<t}, x; \mathcal{R}),
\label{eq:forward_kl}
\end{equation}
where $\text{sg}(\cdot)$ denotes stop-gradient and $\mathcal{M}$ is the set of model-generated token positions (excluding retrieved passages via state masking). When $\bar r_t=r_t$ (no clipping), the expectation under $o_t \sim \pi_\theta$ is gradient-equivalent to minimizing $D_\text{KL}(\pi_\text{tea}(\cdot|o_{<t})\|\pi_\theta(\cdot|o_{<t}))$ averaged over student-sampled prefixes. In implementation, $\epsilon{=}10^{-6}$ and $R_\text{max}{=}10$; clipping introduces controlled bias relative to the exact forward-KL gradient while limiting importance-weight magnitude. The implemented gradient is:
\begin{equation}
\nabla_\theta \mathcal{L}_\text{OPD}^{\text{fwd}} = -\frac{1}{|\mathcal{M}|}\sum_{t\in\mathcal{M}} \text{sg}(\bar r_t) \cdot \nabla_\theta \log\pi_\theta(o_t),
\label{eq:fwd_grad}
\end{equation}
a weighted policy gradient where tokens with $r_t > 1$ (teacher favors more than student) receive amplified gradients up to the cap $R_\text{max}$, encouraging the student to cover the teacher's distribution.

\paragraph{Reverse KL surrogate variant.}
We also experiment with a convex surrogate for the reverse KL objective $D_\text{KL}(\pi_\theta \| \pi_\text{tea})$, using $f(u) = e^u - u - 1$ applied to $u_t = \ell_t - \ell_t^\text{tea}$:
\begin{equation}
\mathcal{L}_\text{OPD}^{\text{rev}}(\theta) = \frac{1}{|\mathcal{M}|}\sum_{t \in \mathcal{M}} \left[ \exp\!\left(\ell_t - \ell_t^\text{tea}\right) - \left(\ell_t - \ell_t^\text{tea}\right) - 1 \right],
\label{eq:reverse_kl}
\end{equation}
where $\ell_t^\text{tea}$ is detached. Unlike forward-KL where high-weight tokens are rarely sampled, this surrogate's high-coefficient tokens have high $\pi_\theta$ and are frequently sampled, making gradients prone to entropy collapse (Appendix G of the supplementary material).

\paragraph{Regularization and masking.}
A KL penalty $\mathcal{L}_\text{KL} = D_\text{KL}(\pi_\theta \| \pi_\text{ref})$ against the frozen initial checkpoint prevents catastrophic deviation. State masking excludes retrieved-passage tokens from all losses.

\subsection{Connection to RL and Objective Choice}
\label{sec:entropy}

OPDSearch+ can be viewed as a form of policy gradient with $R_t = \text{sg}(\bar r_t)$ as a per-token reward derived from the teacher, providing dense, stable supervision at every token (unlike trajectory-level outcome rewards). The trade-off is that this signal reflects teacher preferences rather than task-specific outcomes, motivating the subsequent RL stage. A detailed comparison of forward-KL vs.\ reverse-KL gradient properties is provided in Appendix G of the supplementary material.

% ====================================================================
\section{Experiments}
% ====================================================================

\begin{table*}[t]
\centering
\resizebox{\textwidth}{!}{
\begin{tabular}{l c ccc c cccc c}
\toprule
\multirow{2}{*}{\textbf{Method}}
    & \multirow{2}{*}{\textbf{Size}}
    & \multicolumn{3}{c}{\textbf{Single-Hop QA}}
    & \multirow{2}{*}{\textbf{SH-Avg}}
    & \multicolumn{4}{c}{\textbf{Multi-Hop QA}}
    & \multirow{2}{*}{\textbf{Avg}} \\
\cmidrule(lr){3-5} \cmidrule(lr){7-10}
& & NQ$^\dagger$ & TriviaQA$^\star$ & PopQA$^\star$
& & HotpotQA$^\dagger$ & 2Wiki$^\star$ & Musique$^\star$ & Bamboogle$^\star$ & \\
\midrule
\multicolumn{11}{l}{\textit{3B Methods (same model size as OPDSearch+)}} \\
\midrule
Direct Generation$^\diamond$  & 3B & 0.106 & 0.288 & 0.108 & 0.167 & 0.149 & 0.244 & 0.020 & 0.024 & 0.134 \\
SFT$^\diamond$                & 3B & 0.249 & 0.292 & 0.104 & 0.215 & 0.186 & 0.248 & 0.044 & 0.112 & 0.176 \\
Naive RAG$^\diamond$          & 3B & 0.348 & 0.544 & 0.387 & 0.426 & 0.255 & 0.226 & 0.047 & 0.080 & 0.270 \\
Search-o1$^\diamond$          & 3B & 0.238 & 0.472 & 0.262 & 0.324 & 0.221 & 0.218 & 0.054 & 0.320 & 0.255 \\
Search-R1-Base$^\diamond$     & 3B & 0.421 & 0.583 & 0.413 & 0.472 & 0.297 & 0.274 & 0.066 & 0.128 & 0.312 \\
Search-R1-Instruct$^\diamond$ & 3B & 0.397 & 0.565 & 0.391 & 0.451 & 0.331 & 0.310 & 0.124 & 0.232 & 0.336 \\
ReSearch-Base$^\diamond$      & 3B & 0.427 & 0.597 & 0.430 & 0.485 & 0.305 & 0.272 & 0.074 & 0.128 & 0.319 \\
ReSearch-Instruct$^\diamond$  & 3B & 0.365 & 0.571 & 0.395 & 0.444 & 0.351 & 0.272 & 0.095 & 0.266 & 0.331 \\
AutoRefine-Base$^\diamond$    & 3B & \underline{0.467} & \underline{0.620} & \underline{0.450} & \underline{0.512} & \underline{0.405} & \underline{0.393} & \underline{0.157} & 0.344 & \underline{0.405} \\
AutoRefine-Instruct$^\diamond$& 3B & 0.436 & 0.597 & 0.447 & 0.493 & 0.404 & 0.380 & 0.169 & 0.336 & 0.396 \\
StepSearch-Base$^\diamond$    & 3B & --- & --- & --- & --- & 0.329 & 0.339 & \underline{0.181} & 0.328 & --- \\
StepSearch-Instruct$^\diamond$& 3B & --- & --- & --- & --- & 0.345 & 0.320 & 0.174 & 0.344 & --- \\
GiGPO-Instruct$^\diamond$    & 3B & 0.420 & 0.595 & 0.424 & 0.480 & 0.369 & 0.370 & 0.126 & \textbf{0.641} & 0.421 \\
\midrule
\multicolumn{11}{l}{\textit{Our Method (3B student, 14B-Instruct teacher, OPD $\to$ RL)}} \\
\midrule
\rowcolor{gray!10}
\textbf{OPDSearch+ (Ours)} & \textbf{3B} & \textbf{0.4852} & \textbf{0.6226} & \textbf{0.4513} & \textbf{0.520} & \textbf{0.4580} & \textbf{0.4264} & \textbf{0.1819} & \underline{0.4560} & \textbf{0.4402} \\
\bottomrule
\end{tabular}
}
\caption{Performance comparison on QA benchmarks (Exact Match). OPDSearch+ uses a \textbf{3B} student model. $^\dagger$ indicates in-distribution training data. $^\star$ indicates out-of-distribution test sets. $^\diamond$ indicates numbers cited from prior work. Best in \textbf{bold}, second best \underline{underlined}.}
\label{tab:main_results}
\end{table*}

\subsection{Experimental Setup}

\paragraph{Models and data.}
The student is \textbf{Qwen2.5-3B} (base); the teacher is \textbf{Qwen2.5-14B-Instruct} (frozen). Training uses NQ~\cite{kwiatkowski2019natural} + HotpotQA~\cite{yang2018hotpotqa} train splits ($\sim$170k QA pairs). Evaluation covers seven benchmarks: NQ, TriviaQA~\cite{joshi2017triviaqa}, PopQA~\cite{mallen2023popqa} (single-hop) and HotpotQA, 2WikiMultihopQA~\cite{ho2020constructing}, MuSiQue~\cite{trivedi2022musique}, Bamboogle~\cite{press2023bamboogle} (multi-hop). Retrieval uses E5-base-v2~\cite{wang2024text} over the December 2018 Wikipedia dump~\cite{karpukhin2020dpr} (top-3 passages), identical to Search-R1.

\paragraph{Training configuration.}
Both stages use the veRL framework~\cite{sheng2025verl} on 4$\times$ H200 GPUs with lr $1{\times}10^{-6}$, $G{=}8$ rollouts, $T_\text{max}{=}4$ search turns.
\textit{OPD stage}: trained to the planned horizon (over 420 updates), batch 512, $\alpha_\text{opd}{=}1.0$, forward-KL loss, and no reward signal.
\textit{RL stage}: initialized from the best OPD checkpoint (step 150) and continued until training collapse was observed, at which point the run was stopped; batch 1024, GRPO with $R = 0.9 R_\text{EM} + 0.1 R_\text{format}$, and no teacher. Full hyperparameters in Appendix A of the supplementary material.

\paragraph{Baselines.}
We compare against Search-R1~\cite{jin2025searchr1}, ReSearch~\cite{research}, AutoRefine~\cite{autorefine2025}, StepSearch~\cite{wang2025stepsearch}, GiGPO~\cite{feng2025gigpo}, and non-RL baselines (Direct Generation, SFT, Naive RAG, Search-o1). All use 3B models.

\subsection{Main Results}

Table~\ref{tab:main_results} presents the main results. OPDSearch+ with a \textbf{3B model} achieves a mean EM of $\mathbf{0.4402}$, the best among all 3B methods (surpassing GiGPO-Instruct at $0.421$).

\paragraph{Single-hop QA.}
OPDSearch+ achieves NQ $0.4852$, TriviaQA $0.6226$, PopQA $0.4513$ (SH-Avg $0.520$), exceeding AutoRefine-Base ($0.512$), the best 3B method.

\paragraph{Multi-hop QA.}
The largest gains appear on multi-hop benchmarks: HotpotQA $0.4580$ ($+13.1\%$ over AutoRefine-Base), 2Wiki $0.4264$ ($+8.5\%$ over AutoRefine-Base), MuSiQue $0.1819$ (best among all 3B methods), Bamboogle $0.4560$ (second only to GiGPO's $0.641$ on this 125-sample set).

\paragraph{Key observation: Distillation excels on multi-hop.}
We attribute the disproportionate multi-hop gains to the teacher's multi-step decomposition capabilities: the 14B-Instruct teacher naturally generates sub-queries for complex questions, and on-policy distillation transfers these behaviors efficiently. In contrast, 3B RL models must discover such strategies from scratch through exploration with sparse rewards.

\subsection{Training Dynamics Analysis}
\label{sec:analysis}

\begin{figure}[t]
\centering
\includegraphics[width=\columnwidth]{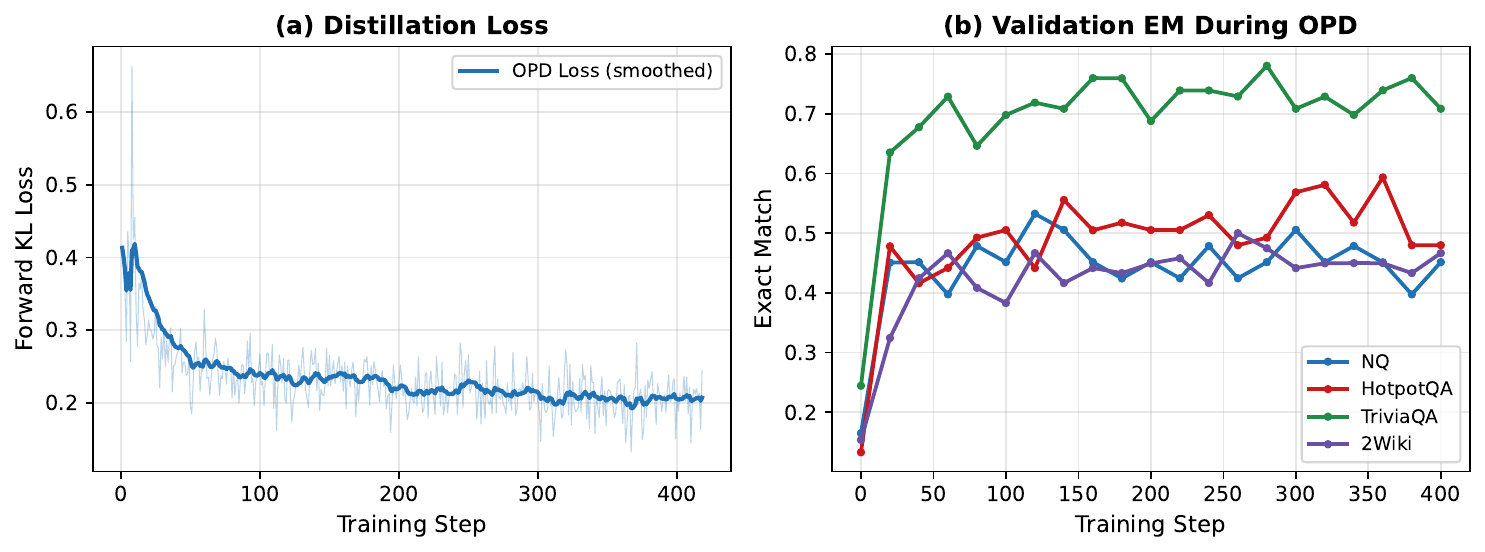}
\caption{OPD training dynamics (14B$\to$3B, forward-KL). (a) Distillation loss decreases rapidly in the first $\sim$50 steps and stabilizes around $0.2$. (b) Validation EM improves progressively across all benchmarks.}
\label{fig:opd_curves}
\end{figure}

Figure~\ref{fig:opd_curves} shows the training dynamics. The distillation loss decreases from $\sim$0.66 to $\sim$0.20, reflecting an emergent curriculum where supervision intensity naturally decreases as the student improves. Query formulation quality improves progressively---by step 200, queries become entity-specific and decomposed rather than verbatim question copies. Training is stable throughout 501 steps without oscillations or divergence, consistent with the variance analysis in Section~\ref{sec:entropy}.

\subsection{Teacher-Weighted CE vs.\ Reverse-KL Surrogate}
\label{sec:kl_comparison}

\begin{table}[t]
\centering
\resizebox{\columnwidth}{!}{%
\begin{tabular}{l l c c}
\toprule
\textbf{Objective} & \textbf{Stability} & \textbf{Best Val EM} & \textbf{Grad Norm} \\
\midrule
Forward-KL (Ours) & Stable (400+ steps) & \textbf{0.493} & 0.73--1.19 \\
Rev-KL Adaptive & Stable (160+ steps) & 0.470 & 0.23--0.32 \\
Rev-KL Clipped & Degrades by step $\sim$60 & 0.136 & 2.2--4.5 \\
Rev-KL JSD & Collapses at step $\sim$20 & --- & 75.0 $\to$ NaN \\
Rev-KL Log-Ratio & Collapses at step $\sim$20 & --- & $\to$ NaN \\
\bottomrule
\end{tabular}}
\caption{Comparison of distillation objectives under matched hyperparameters (all 3B student, 14B teacher, same lr/$\alpha_\text{kl}$/rollout config). Under these shared settings, only the entropy-adaptive variant avoids collapse among reverse-KL alternatives, but it still underperforms our forward-KL objective. Collapses may be mitigable with objective-specific tuning (see text).}
\label{tab:kl_comparison}
\end{table}

\begin{figure}[t]
\centering
\includegraphics[width=\columnwidth]{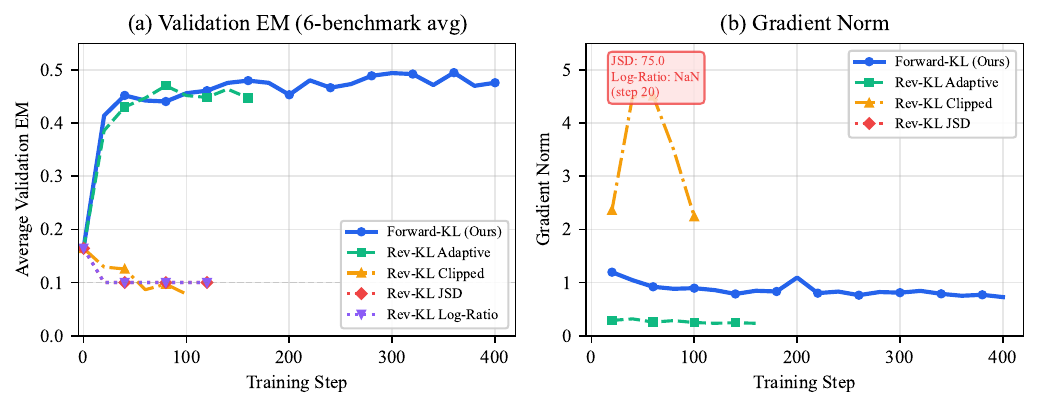}
\caption{Distillation objective comparison under matched hyperparameters (all 14B$\to$3B). (a) Validation EM: Forward-KL (ours) achieves the highest performance and continues improving through 400 steps. Rev-KL Adaptive avoids collapse but plateaus below forward-KL. Three other reverse variants collapse under default settings. (b) Gradient norm: Forward-KL maintains stable norms ($\sim$0.8--1.2); Adaptive has lower norms ($\sim$0.3) but this does not translate to better EM.}
\label{fig:training_stability}
\end{figure}

We compare our forward-KL objective against four reverse-KL-motivated alternatives under identical settings (same teacher, student, lr, KL coefficient; only the loss differs). Results are in Table~\ref{tab:kl_comparison} and Figure~\ref{fig:training_stability}.

Under shared hyperparameters, JSD and Log-Ratio produce NaN gradients within 20 steps, Clipped degrades by step 100, and only the entropy-adaptive variant~\cite{jin2026eopd} avoids collapse. Even when stabilized, the adaptive variant achieves lower EM (0.470 vs.\ 0.493) and saturates by step $\sim$80 while forward-KL continues improving through step 400. This confirms that reverse-KL objectives require substantially more engineering effort and still underperform in the cross-model search setting, consistent with the gradient variance asymmetry analyzed in Appendix G of the supplementary material.

\subsection{Reward Granularity Ablation}
\label{sec:reward_ablation}

We conduct a controlled ablation comparing five RL reward strategies (Pure RL, DAPO token-level, PRIME implicit, Dr.\ GRPO, Turn-level) against OPD (the reward-granularity figure and Appendix H of the supplementary material). Key findings: (1) OPD never collapses (420+ steps), consistent with the clipped second-moment control of Property~2. (2) All outcome-based RL methods eventually collapse. (3) Dense token-level rewards (DAPO, PRIME) are stable but plateau below OPD$\to$RL's peak. (4) Coarse reward shaping (Dr.\ GRPO, Turn-level) fails catastrophically within 40--70 steps.

\begin{table*}[t]
\centering
\resizebox{\textwidth}{!}{
\begin{tabular}{l ccc c cccc c}
\toprule
\multirow{2}{*}{\textbf{Method}}
    & \multicolumn{3}{c}{\textbf{Single-Hop QA}}
    & \multirow{2}{*}{\textbf{SH-Avg}}
    & \multicolumn{4}{c}{\textbf{Multi-Hop QA}}
    & \multirow{2}{*}{\textbf{Avg}} \\
\cmidrule(lr){2-4} \cmidrule(lr){6-9}
& NQ & TriviaQA & PopQA
& & HotpotQA & 2Wiki & Musique & Bamboogle & \\
\midrule
\multicolumn{10}{l}{\textit{Teacher model evaluated directly (no training)}} \\
\midrule
Qwen2.5-7B-Instruct (teacher) & 0.3019 & 0.5633 & 0.3404 & 0.4019 & 0.2811 & 0.2624 & 0.1059 & 0.3040 & 0.3084 \\
Qwen2.5-14B-Instruct (teacher) & 0.3630 & \textbf{0.6479} & 0.3919 & 0.4676 & 0.3916 & 0.3513 & 0.1592 & \textbf{0.4844} & 0.3985 \\
\midrule
\multicolumn{10}{l}{\textit{Pure RL baseline (GRPO, no distillation)}} \\
\midrule
Search-R1-Base (3B)$^\diamond$     & 0.421 & 0.583 & 0.413 & 0.472 & 0.297 & 0.274 & 0.066 & 0.128 & 0.312 \\
Search-R1-Instruct (3B)$^\diamond$ & 0.397 & 0.565 & 0.391 & 0.451 & 0.331 & 0.310 & 0.124 & 0.232 & 0.336 \\
Pure RL (reproduced, 3B)$^\ddagger$ & 0.4660 & 0.6211 & \textbf{0.4540} & 0.5137 & 0.3467 & 0.3142 & 0.0885 & 0.1760 & 0.3524 \\
\midrule
\multicolumn{10}{l}{\textit{Offline SFT (teacher-generated correct trajectories)}} \\
\midrule
Offline SFT (14B $\to$ 3B)    & 0.3613 & 0.5315 & 0.3540 & 0.4156 & 0.2945 & 0.3123 & 0.1133 & 0.3040 & 0.3264 \\
Offline SFT $\to$ RL (14B $\to$ 3B) & 0.4805 & 0.6049 & 0.4293 & 0.5049 & 0.4393 & 0.3887 & 0.1788 & 0.4080 & 0.4185 \\
\midrule
\multicolumn{10}{l}{\textit{Pure OPD (forward-KL distillation only, no RL)}} \\
\midrule
OPD only (14B $\to$ 3B)       & 0.3574 & 0.5872 & 0.3838 & 0.4428 & 0.3602 & 0.3281 & 0.1419 & 0.4000 & 0.3655 \\
OPD only (7B $\to$ 3B)        & 0.3306 & 0.5534 & 0.3518 & 0.4119 & 0.2872 & 0.2875 & 0.1150 & 0.3920 & 0.3311 \\
\midrule
\multicolumn{10}{l}{\textit{Joint: RL + OPD simultaneously (GRPO + forward-KL, opd\_coef=1.0)}} \\
\midrule
Joint RL+OPD (14B $\to$ 3B) & 0.3597 & 0.5908 & 0.3810 & 0.4438 & 0.3736 & 0.3426 & 0.1463 & 0.3680 & 0.3660 \\
\midrule
\multicolumn{10}{l}{\textit{Two-stage: OPD warmup $\to$ GRPO RL fine-tuning}} \\
\midrule
OPD $\to$ RL (7B $\to$ 3B)  & 0.4665 & 0.6057 & 0.4264 & 0.4995 & 0.4357 & 0.3944 & \textbf{0.1910} & 0.4240 & 0.4205 \\
\rowcolor{gray!10}
\textbf{OPD $\to$ RL (14B $\to$ 3B)} & \textbf{0.4852} & 0.6226 & 0.4513 & \textbf{0.5197} & \textbf{0.4580} & \textbf{0.4264} & 0.1819 & 0.4560 & \textbf{0.4402} \\
\bottomrule
\end{tabular}
}
\caption{Ablation study on teacher scale and two-stage training (full test set evaluation, Exact Match). All methods use Qwen2.5-3B as the student/policy model. ``OPD only'' denotes pure on-policy distillation without RL. ``OPD $\to$ RL'' denotes the two-stage approach (OPD warmup followed by GRPO fine-tuning). ``Offline SFT'' trains on teacher-generated correct trajectories (EM-filtered). $^\diamond$ indicates numbers cited from prior work. $^\ddagger$ indicates our reproduced RL baseline using $R = 0.9 \cdot R_\text{EM} + 0.1 \cdot R_\text{format}$ (same reward as our OPD$\to$RL stage). Best in \textbf{bold}.}
\label{tab:ablation_full}
\end{table*}
\subsection{How OPD Reshapes the Policy Distribution}
\label{sec:reshaping}

A central claim of this work is that OPD \emph{reshapes} the student's policy distribution, enabling subsequent RL to converge to solutions unreachable from the base initialization. We provide quantitative evidence (detailed plots in Appendix I of the supplementary material):

\paragraph{OPD expands behavioral diversity.}
The OPD-initialized student enters RL with entropy $1.35$---significantly higher than the base model's $0.82$. OPD does not merely sharpen toward the teacher's preferred actions, but \emph{broadens} the student's action space by transferring diverse search strategies.

\paragraph{Controlled policy drift and search behavior.}
Forward-KL OPD produces gradual, controlled drift (KL $\sim$0.2 from reference), while pure RL drifts erratically (KL $>$0.8 before collapse). OPD also reduces average search turns from $3.5$ to $2.7$, indicating more precise query formulation that outcome-only RL struggles to discover from sparse rewards.

\paragraph{Gradient stability.}
Forward-KL OPD maintains monotonically decreasing gradient norms ($1.1 \to 0.7$ over 418 steps), enabling sustained learning. The RL-post-OPD stage starts with low norms ($\sim$0.3), consistent with the OPD-initialized policy being closer to a good solution.

% ====================================================================
\section{Discussion}
% ====================================================================

\paragraph{Off-the-shelf instruct models as teachers.}
General-purpose instruct models serve as effective OPD teachers because they have internalized information-seeking behaviors through broad instruction-tuning, and the forward-KL objective efficiently extracts these capabilities into the student's on-policy context.

\paragraph{Complementarity with RL.}
OPD and RL contribute complementary capabilities: OPD establishes search behaviors while RL refines answer accuracy. Joint optimization ($0.3660$) substantially underperforms the sequential pipeline ($0.4402$) due to objective interference. The total cost is comparable to Search-R1's RL training ($\sim$48 GPU-hours on 4$\times$ H200).

\subsection{Ablation: Teacher Scale and Two-Stage Training}

Table~\ref{tab:ablation_full} presents the ablation on teacher scale and two-stage training.

\paragraph{Pure RL is weak on multi-hop.}
Our reproduced pure RL baseline achieves competitive single-hop performance (SH-Avg $0.514$) but dramatically underperforms on multi-hop: HotpotQA $0.347$, 2Wiki $0.314$, Bamboogle $0.176$ (Avg $0.352$). OPD$\to$RL achieves HotpotQA $0.458$ ($+32.1\%$), 2Wiki $0.426$ ($+35.7\%$), confirming that OPD provides the multi-step reasoning foundation that pure RL cannot bootstrap from sparse rewards.

\paragraph{OPD outperforms offline SFT.}
Pure OPD ($0.3655$) outperforms Offline SFT ($0.3264$, $+12.0\%$) without trajectory filtering or data construction. Even with RL, Offline SFT$\to$RL ($0.4185$) underperforms OPD$\to$RL ($0.4402$, $+5.2\%$), supporting Proposition~\ref{prop:onpolicy} on distribution shift.

\paragraph{Two-stage training is key.}
OPD$\to$RL ($0.4402$) outperforms pure OPD ($0.3655$, $+20.4\%$), confirming that OPD provides the behavioral foundation while RL refines answer accuracy. Even the 7B-teacher two-stage pipeline ($0.4205$) exceeds pure OPD with the larger 14B teacher. The corresponding training curve is provided in the supplementary material.

% ====================================================================
\section{Conclusion}
% ====================================================================

We present OPDSearch+, an on-policy distillation framework for search-augmented reasoning that uses a frozen off-the-shelf teacher and requires no task-specific training. The two-stage pipeline (OPD $\to$ RL) with a 3B student achieves mean EM $0.4402$, outperforming all 3B baselines with particularly strong multi-hop gains.

directly applying OPD with an off-the-shelf teacher without task-specific fine-tuning
\bibliography{reference}

% ====================================================================

\end{document}